\documentclass[lettersize,journal]{IEEEtran}
\usepackage{amsmath,amsfonts}
\usepackage{algorithmic}
\usepackage{algorithm}
\usepackage{array}
\usepackage[caption=false,font=normalsize,labelfont=sf,textfont=sf]{subfig}
\usepackage{textcomp}
\usepackage{stfloats}
\usepackage{url}
\usepackage{verbatim}
\usepackage{graphicx}
\usepackage{cite}
\usepackage{caption}
\usepackage{diagbox}
\usepackage{tabu}                     
\usepackage{multirow}                 
\usepackage{multicol}                 
\usepackage{float}                    
\usepackage{makecell}                 
\usepackage{booktabs}
\usepackage{bbm}
\usepackage{graphicx}
\usepackage{array}
\usepackage{makecell}
\usepackage{algorithm}
\usepackage{algorithmic}
\usepackage{colortbl}
\usepackage{setspace}
\usepackage{color}
\definecolor{mygray}{gray}{.9}

\newcommand{\eg}{\textit{e}.\textit{g}.}

\newcommand{\ie}{\textit{i}.\textit{e}.}
\newcommand{\etc}{\textit{etc}}

\newcommand{\wrt}{\textit{w.r.t.}}

\begin{document}

\title{Unsupervised Visible-Infrared ReID via Pseudo-label Correction and Modality-level Alignment}

\author{Yexin Liu*, Weiming Zhang*, Athanasios  V. Vasilakos, and Lin Wang$^\dagger$
\thanks{Manuscript received April 1, 2024. $^\dagger$ corresponding author. * co-first author.}
\thanks{Y. Liu and W. Zhang are with the Artificial Intelligence Thrust, HKUST(GZ), Guangzhou, China. E-mail:yliu292@connect.hkust-gz.edu.cn and zweiming996@gmail.com}
\thanks{A. Vasilakos is with the Center for AI Research (CAIR), University of Agder(UiA), Grimstad, Norway. Email: th.vasilakos@gmail.com}
\thanks{L. Wang is with the Artificial Intelligence Thrust, HKUST(GZ), Guangzhou, and Dept. of Computer Science and Engineering, HKUST, Hong Kong SAR, China. E-mail: linwang@ust.hk
}}

\markboth{Journal of \LaTeX\ Class Files,~Vol.~14, No.~8, August~2021}%
{Shell \MakeLowercase{\textit{et al.}}: A Sample Article Using IEEEtran.cls for IEEE Journals}

\maketitle

\begin{abstract}
Unsupervised visible-infrared person re-identification (UVI-ReID) has recently gained great attention due to its potential for enhancing human detection in diverse environments without labeling. Previous methods utilize intra-modality clustering and cross-modality feature matching to achieve UVI-ReID. However, there exist two challenges: 1) noisy pseudo labels might be generated in the clustering process, and 2) the cross-modality feature alignment via matching the marginal distribution of visible and infrared modalities may misalign the different identities from two modalities.
In this paper, we first conduct a theoretic analysis where an interpretable generalization upper bound is introduced.  Based on the analysis, we then propose a novel unsu\textbf{p}ervised c\textbf{r}oss-mod\textbf{a}l\textbf{i}ty per\textbf{s}on r\textbf{e}-identification framework (\textbf{PRAISE}).
Specifically, to address the first challenge, we propose a pseudo-label correction strategy that utilizes a Beta Mixture Model to predict the probability of mis-clustering based network's memory effect and rectifies the correspondence by adding a perceptual term to contrastive learning. Next, we introduce a modality-level alignment strategy that generates paired visible-infrared latent features and reduces the modality gap by aligning the labeling function of visible and infrared features to learn identity discriminative and modality-invariant features.
Experimental results on two benchmark datasets demonstrate that our method achieves state-of-the-art performance than the unsupervised visible-ReID methods.
\end{abstract}


\section{Introduction}
\label{sec:intro}

Person re-identification (ReID) aims to identify individuals across disparate cameras, considering various challenges such as changes in viewpoint, lighting conditions, and partial obstructions. It plays a critical role in numerous applications, notably in security surveillance, intelligent transportation systems, robotics, and smart cities.~\cite{lang2023self,ye2023person,li2021progressive,zhang2022fmcnet,wang2022nformer}.
Numerous methods focus on matching pedestrians with visible cameras and regard ReID as a single-modality learning problem~\cite{xu2018attention,pu2023memorizing,yang2018person}.
However, these methods may not be effective for images captured by visible cameras in poor illumination conditions, \eg, nighttime, making it difficult to extract reliable visual information for identification~\cite{choi2020hi}.
For this reason, infrared cameras, more robust to illumination changes, are introduced to complement the visible cameras~\cite{feng2019learning,wu2021discover,choi2020hi,zheng2022visible}.
Consequently, intensive research endeavors have been made for visible-infrared person re-identification (VI-ReID), which aims to retrieve the target samples in one modality when given a query sample from another modality~\cite{park2021learning,zhang2022fmcnet,cai2021dual}.

The key challenge of VI-ReID is how to bridge both modalities by minimizing the large modality gap between visible and infrared images.
Most VI-ReID methods focus on learning modality-invariant features via disentanglement of the modality-specific semantic and structural information, mutual translation, or metric learning~\cite{wu2017rgb,ye2019bi,lu2020cross,wang2019learning,wang2019rgb}, \etc. Despite their success, they require the availability of well-annotated visible-infrared datasets for supervision. These datasets, however, often require intensive label annotation endeavors, especially for infrared images, which lack color information. For this reason, unsupervised domain adaptation (UDA) methods have been proposed to learn cross-modality models with only one modality (\eg, visible) available for supervision~\cite{liang2021homogeneous,wang2022optimal}.
However, these methods still rely on labeled visible datasets, which may cause privacy concerns and thus limit their real-world application values.

\begin{figure}[t!]
    \centering
    \includegraphics[width=0.5\textwidth]{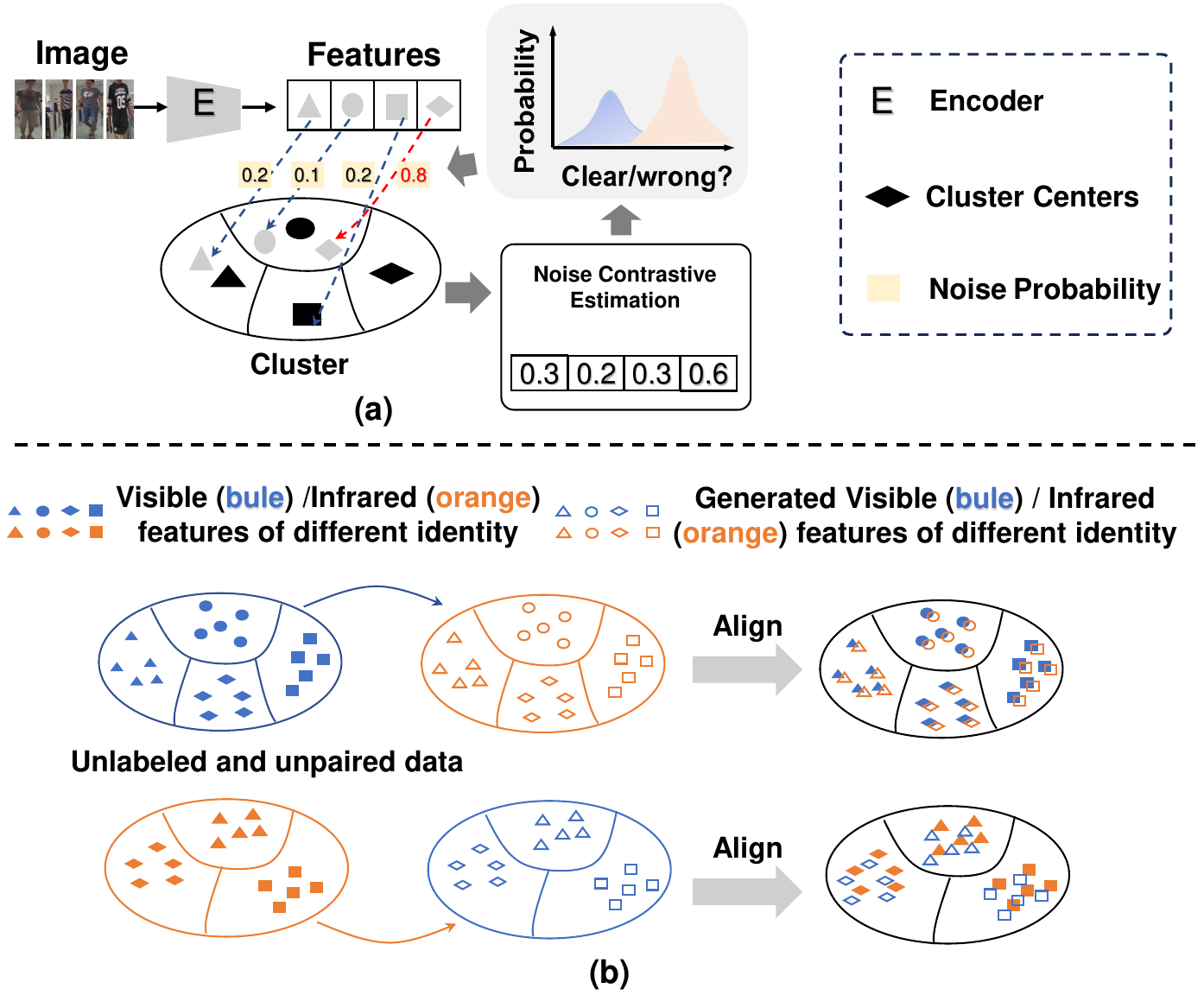}
    \caption{Illustration of the proposed two strategies. (a) pseudo label correction, (b) modality-level alignment.}
    \vspace{-10pt}
    \label{cover figure}
\end{figure}

To address these challenges, some unsupervised VI-ReID methods have been proposed by utilizing intra-modality clustering and cross-modality feature alignment via matching. However, there exist two problems: 1) Applying the clustering algorithms~\cite{liang2021homogeneous,hu2022feature}, to generate pseudo labels may generate noisy pseudo-labels. Directly training the model with noisy samples may degrade its performance. 2) The cross-modality feature alignment, via matching the marginal distribution of visible and infrared modalities, may misalign the different identities between the two modalities. The reason is that 1) Different pedestrians in the target matching modality may have similar appearances, such as similar clothes and body shapes, making it nontrivial to distinguish the appearance differences. 2) The same pedestrian in the target matching modality may present significant intra-class variations in appearance, such as changes in illumination, background clutter, pose, viewpoint, and occlusion~\cite{hu2022feature,wang2023correspondence}.

\textbf{Motivation:} In this paper, we strive to answer a crucial question: \textit{Can learning modality-invariant identity-discriminative features be guaranteed by intra-modal clustering and inter-modal matching? If not, under what conditions is it possible?}
We give a negative answer to the first question based on the above analysis. To answer the second question, we conduct a theoretical analysis and introduce an upper bound as a sufficient condition for learning modality-invariant identity-discriminative features by explicitly considering the difference in the distribution of identity features between visible and infrared modalities. In a nutshell, we need to tackle \textit{two} technical hurdles to achieve reliable unsupervised VI-ReID: \textbf{1}) learning identity-discriminative features without any category label; \textbf{2}) reducing the visible-infrared modality gap without paired data.

Based on the theatrical analysis, we propose a novel unsu\textbf{p}ervised c\textbf{r}oss-mod\textbf{a}l\textbf{i}ty per\textbf{s}on r\textbf{e}-identification framework (\textbf{PRAISE}) to learn discriminative feature representation within each modality and conduct cross-modality feature alignment.
Our method includes two key strategies: pseudo-label correction and modality-level alignment.
For pseudo label correction (Sec.~\ref{Pseudo-Label Correction}), based on the network's memory effect, we leverage a two-component Beta Mixture Model (BMM)~\cite{ma2011bayesian} to model both clean and noisy pseudo labels (Fig.\ref{cover figure} (a)), and then utilize the estimated noise probabilities from the BMM model, and incorporate a perceptual term to implement a dynamically weighted cluster-wise contrastive learning loss. This enables us to deal with noisy pseudo labels and learn more robust discriminative features (Tab.~\ref{tab:Ablation two strategies}).
For modality-level alignment (Sec.~\ref{Modality-level alignment Strategy}), we introduce a bi-directional translation module to translate latent features from one modality to the other. We then employ the generated paired latent features and introduce the cross-modality alignment loss to achieve modality-level alignment for the same identity and reduce modality discrepancy, as depicted in Fig.~\ref{cover figure} (b).


\textbf{Contributions:} In summary, our major contributions are as follows:
(\textbf{I}) We theoretically analyze the problem of existing unsupervised VI-ReID methods and propose a novel unsupervised  VI-ReID framework.
(\textbf{II}) We present a novel pseudo-label correction strategy that utilizes a Beta Mixture Model to estimate the noise probability of each sample. We then incorporate a perceptual term into contrastive learning to learn discriminative features.
(\textbf{III}) We propose a modality-level alignment strategy to generate paired visible-infrared features and a cross-modality alignment loss to reduce the modality gap.

\section{Related Works}
\label{sec:Related Works}

\noindent\textbf{Supervised Visible-Infrared Person ReID.}
It is a challenging task due to the large cross-modality discrepancies~\cite{feng2019learning,wu2021discover,choi2020hi,zheng2022visible,park2021learning,ye2020dynamic}. Existing methods mainly focus on optimization from three directions: 1) representation learning, 2) metric learning, and 3) generative model. For representation learning-based methods, some~\cite{wu2017rgb,ye2019bi,liu2021sfanet,ye2020visible,li2024frequency} learn the shared features between different modalities by observing the shared textural and body shape features in the low-level semantic feature space to achieve better cross-modality alignment. Others \cite{choi2020hi,pu2020dual,kansal2020sdl,wu2021discover} focus on the identity-level features, attempting to disentangle identity (ID) discriminative factors from the ID-exclusive factors from the VI image pairs, to highlight the discriminative features of the same identity and reduce identification errors.
For metric learning-based methods, existing methods focus on applying the diverse loss functions (\eg contrastive loss and triplet loss ) ~\cite{ye2018hierarchical,cai2021dual,hao2019hsme} or different training strategies (\eg collaborative learning)~\cite{ye2021channel,liu2022learning,gao2021mso} to mitigate cross-modality discrepancies.
Generative model-based methods~\cite{kniaz2018thermalgan,wang2019rgb,wang2019learning,choi2020hi,wang2020cross,yang2020cross} mainly rely on the generative adversarial networks (GANs) for cross-modality translation. These approaches facilitate the generation of paired cross-modal data, which contribute to the acquisition of modality-invariant features and the mitigation of modality gaps. However, previous approaches have primarily relied on the utilization of extensively annotated labels or, at the very minimum, supervision from one modality, imposing the burden of labor-intensive annotation processes. To this end, we aim to develop a high-performance model in an unsupervised manner to address the challenges of VI-ReID.

\noindent\textbf{Unsupervised Visible-Infrared Person ReID.}
Existing UDA methods relieve the need for large-scale labeled datasets by employing a fraction of the well-labeled visible dataset as the source domain for pre-training to generate reliable pseudo labels~\cite{liang2021homogeneous,wang2022optimal}.
In particular,~\cite{liang2021homogeneous} proposes a homogeneous-heterogeneous learning approach for VI-ReID that utilizes labeled visible datasets for each modality to generate pseudo labels and align the two modalities by learning modality-invariant feature representations.
\cite{wang2022optimal} introduces a UDA approach to generate reliable pseudo labels for unlabeled visible data. An optimal transport strategy is utilized to assign pseudo labels to infrared images.
However, UDA methods still rely on labeled visible datasets, which can not be regarded as fully unsupervised learning methods.
Therefore, ~\cite{yang2022augmented,wu2023unsupervised,yang2023towards,cheng2023unsupervised,yang2023dual,shi2024multi,shi2024progressive} proposed fully unsupervised VI-ReID methods by conducting 1) intra-modality clustering and 2) cross-modality matching. However, the utilization of clustering algorithms for generating pseudo-labels may result in noisy labels, impacting model performance adversely. Additionally, prior methods for cross-modality matching assume similarity between learned features from different modalities, which isn't always accurate. To address these issues, we propose a pseudo-label correction strategy to estimate the noise probability in pseudo-labels, improving feature discriminative ability. We also introduce a modality-level alignment strategy to reduce the gap between visible and infrared modalities.

\vspace{3pt}
\noindent\textbf{Pseudo-Label Correction.}
Various methods have been proposed to refine pseudo-labels in unsupervised person ReID~\cite{fu2022large,ye2021collaborative}, including 1) mutual learning-based refinement, 2) label propagation-based refinement, and 3) supervised clustering. Mutual learning-based refinement approaches like~\cite{cho2022part} refine pseudo-labels by leveraging ensemble predictions of part features, effectively reducing noise by incorporating label smoothing. Label propagation techniques-based methods~\cite{wu2021mgh} are enhanced by integrating metadata such as spatiotemporal information and data correlations across cameras, contributing to the refinement of pseudo-labels. Supervised clustering methods~\cite{yan2022plug} rectify labels obtained from a separately trained network, augmenting sample features through the aggregation of identifiable features from neighboring samples. However, these methods encounter the accumulation of erroneous labels and increased training time. In this paper, We first determine which cluster center is closer to a sample, predict the noise probability using the network's memorization effect~\cite{arpit2017closer}, and then introduce a regulation term to contrastive learning for alleviating the noising problem.

\vspace{-6pt}
\section{Methods}
\label{sec:Methods}


\subsection{Theoretical Analysis}
\label{Theoretical Analysis}
The establishment of a discrepancy measure between two domains plays a crucial role in quantifying their similarity. ~\cite{ben2006analysis} introduces the ${\cal H}$-divergence as a measure for quantifying the dissimilarity between two distributions. We adopt the ${\cal H}$-divergence to characterize the discrepancy between the visible and infrared modalities. The definition of the ${\cal H}$-divergence is as follows:

\textbf{Definition 1} (${ \mathcal{H}}$-divergence): Let ${\cal H}$ be a hypothesis class defined on the input space ${\cal X}$, and ${\cal A_H}$ be the collection of subsets of ${\cal X}$ that correspond to the support of some hypothesis in ${\cal H}$, denoted as $\mathcal{A}_{\mathcal{H}}:={h^{-1}(1)|h\in\mathcal{H}}$. The ${\cal H}$-divergence between two distributions ${\mathcal{D}}$ and ${\mathcal{D}'}$ is defined as $d_{\mathcal H}(\mathcal D,\mathcal D') :=\sup_{A\in\mathcal{A}_{\mathcal{H}}}|\Pr_{\mathcal{D}}(A)-\Pr_{\mathcal{D}'}(A)|$.

The ${\cal H}$-divergence takes values ranging from 0 to $\infty$. A value of 0 indicates that the two distributions $\mathcal D$ and ${\mathcal{D}'}$ are identical, while larger values indicate increasing dissimilarity between the distributions.

\textbf{Definition 2} (Empirical Rademacher Complexity). ~\cite{zhao2019learning} defined the Empirical Rademacher Complexity of ${\cal H}$ as follows:

\begin{equation}
\begin{aligned}
\operatorname{Rad}_{\mathbf{S}}(\mathcal{H}):=\mathbb{E}_{\boldsymbol{\sigma}}\left[\sup\limits_{h\in\mathcal{H}}\frac{1}{n}\sum\limits_{i=1}^n\sigma_i h(\mathbf{x}_i)\right],
\end{aligned}
\end{equation}

where ${\cal H}$ represents a set of functions mapping from ${\cal X}$ to the interval $[a,b]$. $\textbf{S}=\{\textbf{x}_i\}_{i=1}^n$ refers to a fixed sample of size $n$ with elements in ${\cal X}$, and $\boldsymbol{\sigma}=\{\sigma_i\}_{i=1}^{n}$ denotes a sequence of independent uniform random variables in $\{+1,-1\}$.

\textbf{Theorem 1}
Let $\langle\mathcal{D}_S,f_S\rangle$ and $\langle\mathcal{D}_T,f_T\rangle$ represent the source and target domains, respectively. For any function class $\mathcal{H}\subseteq[0,1]^{\mathcal{X}}$ over $\mathcal{X}$, and $\forall h\in\mathcal{H}$, the following inequality holds:
\begin{equation}
\begin{aligned}
\varepsilon_{T}(h)\leq& \widehat{\varepsilon}_{S}(h)+\widehat{\varepsilon}_{V}(h)+d_{\tilde{\mathcal{H}}}(\widehat{\mathcal{D}}_{S},\widehat{\mathcal{D}}_{T})\\
&+2\mathop{\mathrm{Rad}}_{\mathbf{S}}(\mathcal{H})+4\mathop{\mathrm{Rad}}_{\mathbf{S}}(\tilde{\mathcal{H}})  \\
&+\min\{\mathbb{E}_{\mathcal{D}_S}[|f_T-f_S|],\mathbb{E}_{\mathcal{D}_T}[|f_S-f_T|]\}\\
& + O\left(\sqrt{\frac{\log(1/\delta)}{n}}\right),
  \label{Theorem 1}
\end{aligned}
\end{equation}

where $\mathcal{H}:=\{\operatorname{sgn}(|h(\mathbf{x})-h^{\prime}(\mathbf{x})|-t)|h,h^{\prime}\in|\mathcal{H},0\leq t\leq1\}$. $f_S:\mathcal{X}\to[0,1] \text{and} f_T:\mathcal{X}\to[0,1]$ are the optimal labeling functions on the source and target domains, respectively.
Motivated by this theory, we put forward a generalization upper bound for fully unsupervised visible-infrared person re-identification as follows:

\textbf{Our idea:} Based on ${\cal H}$-divergence and the Empirical Rademacher Complexity~\cite{zhao2019learning},
we derive a \textit{generalization upper bound}, which is given as follows.
Let $\langle\mathcal{V},f_V\rangle$ and $\langle\mathcal{I},f_I\rangle$ be the visible and infrared modalities, respectively. $\widehat{\mathcal{D}}_{V}$ and $\widehat{\mathcal{D}}_{I}$ denote the empirical distributions of the visible and infrared features, generated from the extracted features $\boldsymbol{Z}^{V}$ and $\boldsymbol{Z}^{IR}$, respectively. For any function $\mathcal{H}\subseteq[0,1]^{\mathcal{X}}$ over $\mathcal{X}$, and $\forall
 h\in\mathcal{H}$, the following inequality holds.
\begin{equation}
\begin{aligned}
\varepsilon_{V}(h)+\varepsilon_{I}(h)\leq& \widehat{\varepsilon}_{I}(h)+\widehat{\varepsilon}_{V}(h)+2d_{\tilde{\mathcal{H}}}(\widehat{\mathcal{D}}_{V},\widehat{\mathcal{D}}_{I})\\
&+2(\mathop{\mathrm{Rad}}_{\mathbf{V}}(\mathcal{H})+ \mathop{\mathrm{Rad}}_{\mathbf{I}}(\mathcal{H}))\\
&+2(\mathop{\mathrm{Rad}}_{\mathbf{V}}(\tilde{\mathcal{H}})+\mathop{\mathrm{Rad}}_{\mathbf{I}}(\tilde{\mathcal{H}}))  \\
&+2\min\{\mathbb{E}_{\mathcal{D}_V}[|f_I-f_V|],\mathbb{E}_{\mathcal{D}_I}[|f_I-f_V|]\}\\
& + O\left(\sqrt{\frac{\log(1/\delta)}{n}}\right),
  \label{Theorem 1}
\end{aligned}
\end{equation}
Here, the first term represents the error in the visible modality, while the second term represents the error in the infrared modality. The third term quantifies the discrepancy between the marginal distributions of the two modalities. The fourth and fifth terms correspond to complexity measures of the hypothesis spaces $\mathcal{H}$ and $\tilde{\mathcal{H}}$, respectively. The sixth term describes the shift or difference between the labeling functions of the visible and infrared modalities. The last term captures the error introduced due to the finite size of the sample used for training and testing.
We aim to 1) optimize the first and second items considering the noise label problem and 2) simultaneously perform modal matching and keep consistency between the label functions of visible and infrared modalities.
\begin{figure*}[t!]
    \centering
    \includegraphics[width=0.94\textwidth]{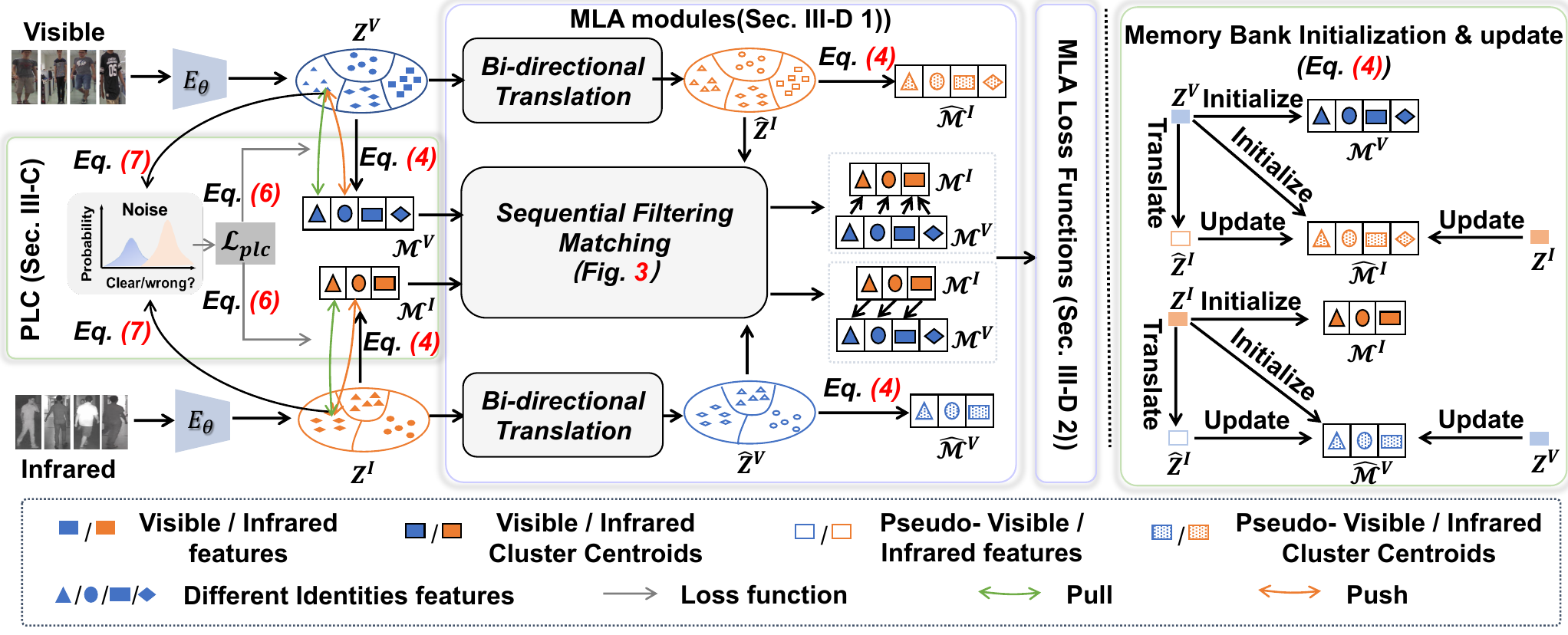}
    \caption{\textbf{Overview of our method}. The encoder extracts the features of visible and infrared images. Then, the extracted features in one modality are translated into another modality. Our method contains two key strategies: pseudo-label correction (PLC) and modality-level alignment (MLA). PLC enables us to deal with noisy pseudo labels and learn more robust discriminative feature. MLA enables us to reduce modality discrepancies.}
    \vspace{-10pt}
    \label{framework}
\end{figure*}
\subsection{Overall Architecture}
\label{Overall Architecture}
Our proposed framework is depicted in Fig.~\ref{framework}. Given the visible data $V$ and infrared data $I$, we denote the $N$ and $S$ as the total numbers of visible and infrared images, respectively. PRAISE aims to encourage the encoder $E_\theta(x)$ to extract identity-discriminative features and reduce the modality gap without any modal supervision.
Based on the above theoretical analysis (See Sec.~\ref{Theoretical Analysis}), we aim to 1) reduce the intra-modality clustering errors. 2) reduce the modality discrepancy during alignment and ensure the consistency of the labeling functions of the visible and infrared modalities. To this end, we propose our framework (named \textbf{PRAISE}) based on two strategies: 1) Pseudo-Label Correction (PLC) (Sec.~\ref{Pseudo-Label Correction}) and 2) Modality-level Alignment (MLA) (Sec.~\ref{Modality-level alignment Strategy}). We now describe these two strategies in detail.

\vspace{-5pt}
\subsection{Pseudo Label Correction (PLC) Strategy}
\label{Pseudo-Label Correction}
Following recent methods~\cite{yang2022augmented,wu2023unsupervised}, we first conduct channel augmentation to the visible images and then utilize the encoder $E_\theta(x)$ to extract the visible and infrared features ($\boldsymbol{Z}^{V}$ and $\boldsymbol{Z}^{I}$). Then, we utilize
DBSCAN method~\cite{khan2014dbscan} to cluster these features in each modality. Based on the clustering results, visible $\mathcal{M}^V$ and infrared $\mathcal{M}^I$ memory banks $\mathcal{M}$ are constructed and initialized by the mean vectors of all the features within each cluster, and then updated by:

\begin{equation}
    \begin{aligned}
            \mathcal{M}^V[j]\leftarrow\lambda\mathcal{M}^V[j]+(1-\lambda)E_\theta(\boldsymbol{Z}^{V}),\\
    \mathcal{M}^I[j]\leftarrow\lambda\mathcal{M}^I[j]+(1-\lambda)E_\theta(\boldsymbol{Z}^{I}),
    \end{aligned}
\end{equation}
where $\mathcal{M}^V[j]$ and $\mathcal{M}^I[j]$ represent the centroids of the visible and infrared memory banks. $\lambda$ is their update rate.

When obtaining the cluster centroids for each modality, previous methods~\cite{yang2022augmented,wu2023unsupervised} directly utilize the ClusterNCE~\cite {dai2022cluster} loss to minimize the distance between the samples within the same cluster and maximize the feature distance between samples belonging to different clusters for each modality (including \textit{visible, infrared, and augmented visible images}). The ClusterNCE loss is formulated as follows:

\begin{equation}
\begin{aligned}
\mathcal{L}_{nce}^i=-\log\frac{exp(\phi_+^{\mathcal{M}}\cdot E_\theta (x_i))}{\sum_{k=1}^K exp({\phi_k^{\mathcal{M}}}\cdot E_\theta (x_i)/\tau)},
  \label{eq:CWC Loss}
\end{aligned}
\end{equation}
where $E_\theta (x_i)$ denotes the extracted feature from $i$-th sample. $\phi_+^{\mathcal{M}}$ and $\phi_k^{\mathcal{M}}$ denote the positive cluster centroid \wrt ~${x_i}$ and $k$-th cluster centeroid, respectively. $\tau$ denotes the temperature parameter. However, the ClusterNCE loss compels features to be closer to the corresponding cluster centroids and ignores that many of these centroids may not be correct. Thus, training the model directly with these incorrect correspondences may degrade the overall model performance.

\textbf{PLC Loss.}
Due to clustering errors, it's common for a cluster to contain noisy features with incorrect pseudo labels. Directly reducing the distance between these noisy features and the cluster centroid is counterproductive for acquiring identity-discriminative features. To address this challenge, we propose a PLC loss to improve the contrastive learning mechanism.
Concretely, our PLC loss involves utilizing the ClusterNCE loss while introducing a perceptual term to harmonize noise pseudo labels based on the model's current state. The formulation is as follows:
\begin{equation}
\begin{aligned}
pull=&(1-w_i) exp(({z_i}\cdot\phi_+)/\tau)+w_i exp(({z_i}\cdot c_i)/\tau)\\
push=&(1-w_i) exp(({z_i}\cdot\phi_+)/\tau)+w_i exp(({z_i}\cdot c_i)/\tau)\\
&+\sum_{k=1,j \neq i}^K exp((z_i\cdot\phi_k)/\tau)\\
&\small\mathcal{L}_{plc}=-\frac{1}{2B}\sum\limits_{i=1}^B (\log\frac{pull}{push}),
  \label{eq:CWC Loss}
\end{aligned}
\end{equation}
where $c_i$ denotes the cluster centroid closest to  $i$-th sample in the current state, and $B$ denotes the batch size.

Based on whether the weights ($w_i$) are dynamically updated, our PLC loss is divided into two types. The first type is called the \textit{Static PLC} (S-PLC) loss, where $w_i$ is set to a constant.
The second type is \textit{Dynamic PLC} (D-PLC) loss,  which involves conducting contrastive learning with dynamic weights $w_i$. Specifically, convolutional neural networks tend to prioritize fitting simple data before learning more complex and noisy examples due to the memorization effects~\cite{arpit2017closer}. The loss functions can indicate whether a sample is clean or noisy~\cite{li2020dividemix}. This allows us to quantitatively assess the likelihood $p$ of a given sample being classified as either clean or noisy, which can be formulated as:
\begin{equation}
\begin{aligned}
p(k\mid\mathcal{L}_{plc}(i)) = \frac{p(k)p(\mathcal{L}_{plc}(i)\mid k)}{p(\mathcal{L}_{plc}(i))},
  \label{eq:CWC Loss}
\end{aligned}
\end{equation}
where clean and noisy pseudo labels are indicated by $k$ = 0 and 1, respectively. The weight $w_i$ is dynamically set to the posterior probability $p(k\mid\mathcal{L}_{plc}(i))$ of the $i$-th sample. Clean samples, characterized by a low noisy probability (\ie, when $1-w_i$ is large), rely more on their pseudo label $\phi+$ for guidance. Alternatively, noisy samples, indicated by a high noisy probability (\ie, when $w_i$ is large), allow their loss to be dominated by their closest cluster centroid $c_i$. To estimate the probability distribution of the loss for a mixture of clean and noisy samples, we employ the Beta Mixture Model (BMM)~\cite{arazo2019unsupervised} as an approximation technique. We train the BMM model with the expectation-maximization algorithm~\cite{moon1996expectation}.
Incorporating the weighted dynamic PLC loss can improve its effectiveness in handling noisy pseudo labels.
Consequently, our proposed PLC loss enhances the overall learning process and improves the model's robustness in handling clustering errors.

\subsection{Modality-level Alignment (MLA) Strategy}
\label{Modality-level alignment Strategy}

\subsubsection{\textbf{MLA modules}}

In VI-ReID, the model is tasked with identifying the same identity's infrared/visible image based on a visible/infrared image. However, the large modality gap presents significant challenges in aligning the two domains. To reduce the modality gap, we first utilize the Bi-directional Translation (BiT) module to generate cross-modality paired latent features.
Then, we propose a sequential filtering matching (SFM) module to obtain the cross-modality cluster centroids correspondence. We now describe these two modules in detail.

\noindent\textbf{Bi-directional Feature Translation (BiT)}.
Aligning unpaired cross-modality images is challenging due to the substantial modality gap.  A straightforward solution is to apply MMD~\cite{liang2021homogeneous} or adversarial learning~\cite{wang2022optimal}.
However, these methods often yield suboptimal performance, as they primarily focus on aligning modality distributions while neglecting the crucial identity alignment (as explained in Sec.~\ref{Theoretical Analysis}). To overcome this challenge, we introduce a Bi-directional Translation module based on CycleGAN~\cite{zhu2017unpaired} to generate the latent features of the other modalities ($\boldsymbol{\hat{Z}}^{I}$or $\boldsymbol{\hat{Z}}^{V}$) utilizing the current latent feature($\boldsymbol{Z}^{V}$or $\boldsymbol{Z}^{I}$), as shown in Fig.~\ref{framework}. We deploy two discriminators to differentiate real and generated latent features and utilize two cycle-consistency losses ($\mathcal{L}_{\mathrm{cyc}}^{\mathrm{V}}$ and
$\mathcal{L}_{\mathrm{cyc}}^{\mathrm{I}}$) to constrain the generator from producing ambiguous features for both visible and infrared modalities. The total loss function for optimizing two generators can be formulated as:
\begin{equation}
\begin{aligned}
\begin{gathered}
\mathcal{L}_{\mathrm{cyc}}^{\mathrm{V}}
=\mathbb{E}_{\boldsymbol{Z}_{i}^V\sim\boldsymbol{Z}^V}\left[\|\boldsymbol{Z}_{i}^V-G_{IV}((\boldsymbol{\hat{Z}}_{i}^V))\|_1\right], \\
\mathcal{L}_{\mathrm{cyc}}^{\mathrm{I}}
=\mathbb{E}_{\boldsymbol{Z}_{i}^{I}\sim\boldsymbol{Z}^{I}}\left[\|\boldsymbol{Z}_{i}^{I}-G_{VI}((\boldsymbol{\hat{Z}}_{i}^{I}))\|_1\right],\\
\mathcal{L}_{GAN} = \mathcal{L}_{cyc}^{V}+\mathcal{L}_{cyc}^{I}+\mathcal{L}_{D_{I}}(\boldsymbol{\hat{Z}}^V)+\mathcal{L}_{D_{V}}(\boldsymbol{\hat{Z}}^{I})
\end{gathered}
  \label{eq:BT Loss}
\end{aligned}
\end{equation}
where $G_{VI}$ and $G_{IV}$ refer to the generators responsible for generating pseudo infrared and visible features, respectively. The $\mathcal{L}_{D_{I}}$ and  $\mathcal{L}_{D_{V}}$ denote the discriminator loss for infrared and visible modalities, respectively.

To optimize the discriminators, we choose the loss from WGAN~\cite{arjovsky2017wasserstein}, known for its improved stability and training dynamics. Therefore, the discriminator loss is written as:
\begin{equation}
\begin{aligned}
\mathcal{L}_{DIS} = D_{V}(\boldsymbol{Z}^{V})-D_{V}(G_{I V}(\boldsymbol{Z}^{IR}))\\
+D_{I}(\boldsymbol{Z}^{IR})-D_{I}(G_{V I}(\boldsymbol{Z}^{V})),
  \label{eq:DIS Loss}
\end{aligned}
\end{equation}
where $D_{V}$ and $D_{I}$ are discriminators for each modality. The BiT module generates paired cross-modal latent features, which serve as a crucial intermediary representation that bridges the gap between modalities.

\noindent\textbf{Sequential Filtering Matching (SFM)}.
\label{Sequential filtering matching}
\begin{figure}[t!]
    \centering
    \includegraphics[width=0.48\textwidth]{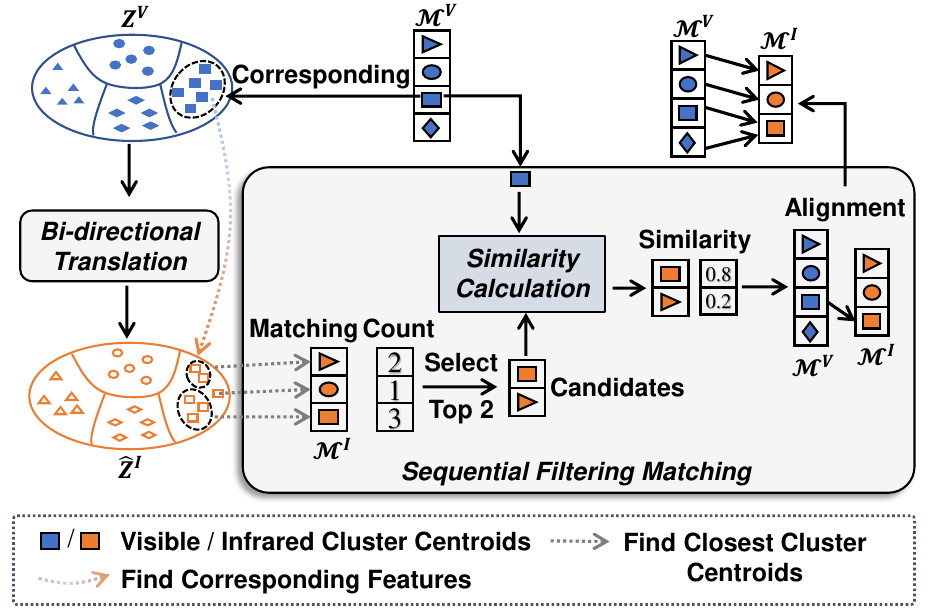}
    \caption{The structure of the proposed sequential filtering matching.}
    \label{Sequential filtering matching1}
    \vspace{-10pt}
\end{figure}
Upon acquiring paired cross-modal latent features, our objective is to deduce the corresponding relationships between real visible cluster centroids ($\mathcal{M}^V$) and infrared cluster centroids ($\mathcal{M}^I$) by evaluating the distribution of the generated pseudo visible/infrared features ($\boldsymbol{\hat{Z}}^{I}$ and $\boldsymbol{\hat{Z}}^{V}$) and to minimize the distance between cross-modal cluster features of matching identities. Specifically, we proposed a sequential filtering matching module to achieve reliable matching between visible and infrared modalities. The SFM module performs matching based on the visible features $\boldsymbol{{Z}}^{V}$ with pseudo infrared features $\boldsymbol{\hat{Z}}^{I}$, and infrared features $\boldsymbol{{Z}}^{I}$ with pseudo visible features $\boldsymbol{\hat{Z}}^{V}$. Taking the case of using visible features and pseudo infrared features as inputs as an example, the specific figure is shown in Fig.~\ref{Sequential filtering matching1}.

Initially, for the input visible features $\boldsymbol{{Z}}^{V}$ and the visible memory bank $\mathcal{M}^V$, we identify the pseudo infrared features $\boldsymbol{\hat{Z}}^{I}$ generated through translation for each visible feature $\boldsymbol{{Z}}^{V}$ within a visible cluster. Subsequently, we measure the distance between these pseudo infrared features $\boldsymbol{\hat{Z}}^{I}$ and the infrared cluster centroids $\mathcal{M}^I$ to determine their respective affiliations. Following this, we select the two infrared cluster centroids with the highest number of pseudo-infrared features as the matching candidates for the current corresponding visible cluster. Finally, we employ a similarity calculation to select the infrared cluster centroid that demonstrates the highest similarity to the currently visible cluster centroid as the result of the matching process. Through the SFM module, we ultimately achieve bi-directional cross-modal cluster centroids matching results, from visible cluster centroids to infrared cluster centroids and vice versa. 

\subsubsection{\textbf{MLA Loss Functions}}
After obtaining the matching results, we introduce the cross-modality alignment (CMA) and Labeling function consistency (LFC) loss to reduce the modality discrepancy and constrain the labeling function of visible and infrared modality to be consistent, as shown in Fig.~\ref{framework}. Specifically, we construct two extra memory banks $\hat{\mathcal{M}}^{V}$ and $\hat{\mathcal{M}}^{I}$ for the pseudo visible and infrared cluster centroids. The $\hat{\mathcal{M}}^{I}$ is initialized by the visible features and updated by the pseudo-infrared features and infrared features. The $\hat{\mathcal{M}}^{V}$ is initialized by the infrared features and updated by the pseudo-visible features and visible features.
\begin{figure}[t!]
    \centering
    \includegraphics[width=0.48\textwidth]{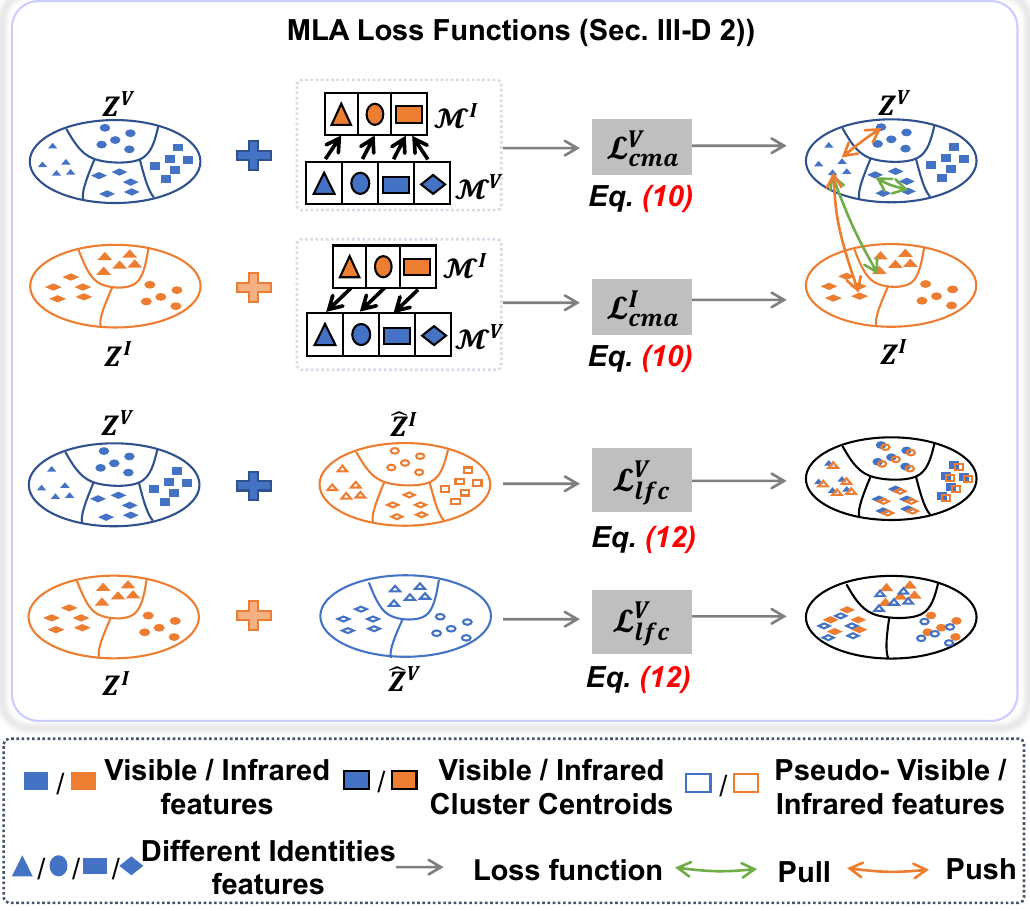}
    \caption{The illustration of the proposed MLA losses. We apply $\mathcal{L}_{cma}$ for cross-modal matching based on feature matching results and clustering centers. Additionally, $\mathcal{L}_{lfc}$ is used to prevent incorrect grouping of different human features from the two modalities.}
    \label{loss}
    \vspace{-12pt}
\end{figure}

\noindent\textbf{Cross-modality Alignment (CMA) loss}.
We propose the CMA loss to leverage the bi-directional cross-modal cluster centroids matching results for modality-level alignment. Using the matching results from visible cluster centroids to infrared cluster centroids as an example: for features from the same visible cluster, we utilize the CMA loss to minimize their distance to the corresponding infrared cluster centroid while maximizing the distance to other infrared cluster centroids. The specific CMA loss $\mathcal{L}_{cma}^V$ for visible features and the total CMA loss are as follows:

\begin{equation}
\begin{gathered}
\mathcal{L}_{cma}^V=-\frac{1}{N}\sum_{i=1}^{N}\log\frac{exp(({\boldsymbol{Z}_i^V}\cdot\phi_+^{{\mathcal{M}}^{I}}/)\tau)}{\sum_{k=1}^K exp(({\boldsymbol{Z}_i^V}\cdot\phi_k^{{\mathcal{M}}^{I}}/)\tau)}, \\
\mathcal{L}_{cma}=\mathcal{L}_{cma}^V+\mathcal{L}_{cma}^{I}.
\end{gathered}
\end{equation}
Although they contain the same inaccurate category information at the beginning, the feature distribution of these cluster centroids will align with that of visible features, and then reducing the modality discrepancy.
\noindent\textbf{Labeling function consistency (LFC) loss}.
\label{Labeling function consistency loss}
For each feature $\boldsymbol{Z}_i^{V}$ and $\boldsymbol{Z}_i^{I}$, we introduce the similarity distribution probability of features that belong to the $k$-th cluster centroid of memory bank $\mathcal{M}$. Take visible features as an example:
\begin{equation}
\begin{aligned}
P(\phi_k^{\mathcal{M}}|\boldsymbol{Z}_i^V)=\frac{exp(({\boldsymbol{Z}_i^V}\cdot\phi_+^{\mathcal{M}}/)\tau)}{\sum_{k=1}^K exp(({\boldsymbol{Z}_i^V}\cdot\phi_k^{\mathcal{M}}/)\tau)},
  \label{eq:CMA Loss_1}
\end{aligned}
\end{equation}

Then, for a given visible features $\boldsymbol{Z}_{i}^{V}$, we define the feature similarity (${F}_{\mathrm{clu}}$) as the positive and negative symmetric cross-entropy of the distribution of probability across the visible and pseudo infrared modality clustering centers. We introduce the LFC loss to further reduce the modality gap by ensuring the labeling functions for visible and infrared features converge toward consistency. Taking visible features $\boldsymbol{{Z}}^{V}$ as an example: for each cluster of visible features $\boldsymbol{{Z}}^{V}$, we aim to minimize their distance to the corresponding pseudo infrared centroid $\phi_+^{\hat{\mathcal{M}}^{I}}$ and maximize their distance to other pseudo infrared centroids $\phi_k^{\hat{\mathcal{M}}^{I}}$.The LFC loss is as follows:

\begin{equation}
\begin{aligned}
& {F}_{\mathrm{clu}}(\boldsymbol{Z}_{i}^{V})=\sum P(\phi_k^{{\mathcal{M}}^{V}}|\boldsymbol{Z}_i^V)\log P(\phi_k^{{\hat{\mathcal{M}}}^{V}}|\boldsymbol{Z}_i^V)\\
& \quad +\sum P(\phi_k^{{\hat{\mathcal{M}}}^{V}}|\boldsymbol{Z}_i^V)\log P(\phi_k^{{\mathcal{M}}^{V}}|\boldsymbol{Z}_i^V)\\
& \mathcal{L}_{lfc}^V=-\frac{1}{N}\sum_{i=1}^{N}\log\frac{e^{{F}_{\mathrm{clu}}(\boldsymbol{Z}_{i}^{V})}}{\sum_{i=1}^{N}e^{{F}_{\mathrm{clu}}(\boldsymbol{Z}_{i}^{V})}}\\
& \mathcal{L}_{lfc}=\mathcal{L}_{lfc}^V+\mathcal{L}_{lfc}^{I}.
\end{aligned}
\end{equation}

By employing the LFC loss, the discrepancy between the labeling functions of visible and infrared features can be minimized. This improves the performance of the model's recognition capabilities.

\section{Experiment}
\label{Experiments}
\subsection{Datasets and implementation details.}

\begin{table*}[t!]
\renewcommand{\arraystretch}{1.15}
\caption{Comparison of different methods on SYSU-MM01 (single-shot) dataset.}
\captionsetup{font=small}
\centering
\resizebox{0.99\textwidth}{!}{
\begin{tabular}{c|c||ccccc||ccccc}
\toprule
\multicolumn{2}{c||}{{Settings}} &\multicolumn{5}{c||}{All Search} & \multicolumn{5}{c}{Indoor Search}  \\ \midrule

Type & Method & Rank-1 & Rank-10 & Rank-20 & mAP & mINP & Rank-1 & Rank-10 & Rank-20 & mAP & mINP\\
 \hline

\multirow{8}{*}{SVI-ReID} & Hi-CMD~\cite{choi2020hi} & 34.9 &77.60& - &35.9& -& -& - &-& -& -\\
& DDAG~\cite{ye2020dynamic} & 54.75& 90.39& 95.81 &53.02 &39.62 &61.02& 94.06& 98.41& 67.98& 62.21\\
& AGW~\cite{ye2021deep} & 47.5 &84.39 &92.14 &47.65 &35.3 &54.17 &91.94 &95.98 &62.97 &59.23\\
& LbA~\cite{park2021learning} & 55.41 &- &- &54.14& - &58.46& - &- &66.33 &-\\
& CAJ~\cite{ye2021channel} & 69.88 & 95.71 & 98.46 & 66.89 & 53.61 & 76.26 & 97.88 & 99.49 & 80.37 & 76.79\\
& MPANet~\cite{wu2021discover} & 70.58 & 96.21 & 98.80 & 68.24 & - & 76.74 & 98.21 & 99.57 & 80.95 & -\\
& FMCNet~\cite{zhang2022fmcnet} & 66.34 & - & - & 62.51 & - & 68.15 & - & - & 74.09 & -\\
& DART~\cite{yang2022learning} & 68.72 &96.36 &98.96 &66.29 &53.26 &72.52& 97.84 &99.46 &78.17 &74.94\\
 \hline
 \hline

 \multirow{5}{*}{USL-ReID} &  SPCL ~\cite{ge2020self} & 18.37 &54.08 &69.02 &19.39 &10.99 &26.83 &68.31& 83.24& 36.42 &33.05 \\
& MMT~\cite{ge2020mutual}  & 21.47 &59.65 &73.29 &21.53 &11.50 &22.79 &63.18 &79.04 &31.50 &27.66\\
& ICE~\cite{chen2021ice}& 20.54 & 57.50 & 70.89&  20.39 & 10.24 & 29.81 & 69.41&  82.66 & 38.35 & 34.32 \\
& IICS~\cite{xuan2021intra}& 14.39 &47.91 &62.32 &15.74& 8.41 &15.91& 54.20 &71.49 &24.87 &22.15\\
& PPLR$^\dagger$~\cite{cho2022part} & 11.98& 43.17 &59.02 &12.25 &4.97 &12.71 &48.66 &68.76 &20.81 &17.61\\
 \hline
 \hline

\multirow{5}{*}{USL-VI-ReID} &  H2H~\cite{liang2021homogeneous} & 30.15 &65.92 &77.32 &29.40 &- &- &- &- &- &-\\
& OTLA~\cite{wang2022optimal}& 29.9 &- &- &27.1& - &29.8& - &- &38.8 &-\\
& ADCA~\cite{lin2019bottom} & 45.51&  85.29 & 93.16 & 42.73 & 28.29 & 50.60 & 89.66 & 96.15 & 59.11 & 55.17\\
& PGMAL~\cite{wu2023unsupervised} & 57.27 & 92.48 & 97.23 & 51.78 & 34.96 & 56.23&  90.19 & 95.39 & 62.74 & 58.13\\
& PRAISE (\textbf{Ours}) & \textbf{59.44}	&\textbf{93.78}	& \textbf{97.62	}& \textbf{53.27}	& \textbf{36.95}	& \textbf{61.03}	& \textbf{94.31}	& \textbf{98.26} & \textbf{66.35} & \textbf{62.21}\\
 \hline
\end{tabular}
}
\label{tab:SYSU Comparison}
\end{table*}

\begin{table}[t!]
\caption{Comparison of different methods on RegDB dataset. }
\renewcommand{\arraystretch}{1.15}
\captionsetup{font=small}
\centering
\resizebox{0.5\textwidth}{!}{
\begin{tabular}{c|c||cc||cc}
\toprule
\multicolumn{2}{c||}{\multirow{1}{*}{Settings}} & \multicolumn{2}{c||}{Visible2Thermal} & \multicolumn{2}{c}{Thermal2Visible} \\ \midrule

Type & Method & Rank-1 & mAP & Rank-1 & mAP\\
 \hline

\multirow{8}{*}{SVI-ReID} & Hi-CMD~\cite{choi2020hi} & 70.93 &66.04 &- &-\\
& DDAG~\cite{ye2020dynamic} & 69.34& 63.46& 68.06 &61.80\\
& AGW~\cite{ye2021deep} & 70.05 &66.37& 70.49& 65.90\\
& LbA~\cite{park2021learning} & 74.17 &67.64& 72.43& 65.46\\
& CAJ~\cite{ye2021channel} & 85.03& 79.14 &84.75& 77.82\\
& MPANet~\cite{wu2021discover} & 83.70& 80.90& 82.80& 80.70\\
& FMCNet~\cite{zhang2022fmcnet} & 89.12& 84.43& 88.38& 83.86\\
& DART~\cite{yang2022learning} & 83.60 &75.67& 81.97& 73.38\\
 \hline
  \hline

 \multirow{6}{*}{USL-ReID} &  SPCL~\cite{ge2020self} & 13.59 &14.68 &11.70 &13.56\\
&MMT~\cite{ge2020mutual} & 25.68& 26.51& 24.42& 25.59\\
&ICE~\cite{chen2021ice} & 12.98 &15.64& 12.18& 14.82\\
&IICS~\cite{xuan2021intra} & 9.17& 9.94& 9.11 &9.90\\
&PPLR~\cite{cho2022part} & 10.30 & 11.94& 10.39 &11.23\\
 \hline
 \hline

\multirow{5}{*}{USL-VI-ReID} &  H2H~\cite{liang2021homogeneous} & 23.81 &18.87&- &-\\
& OTLA~\cite{wang2022optimal}& 32.90& 29.70& 32.10 &28.60\\
&  ADCA~\cite{lin2019bottom} & 67.20 &64.05 &68.48 &63.81 \\
& PGMAL~\cite{wu2023unsupervised} & 69.48 & 65.41 & 69.85 & 65.17\\
& \textbf{PRAISE (Ours)} & \textbf{72.54} & \textbf{68.46} & \textbf{73.15} & \textbf{69.85}\\
 \hline
\end{tabular}
}
\label{tab:RegDB Comparison}
\end{table}

\textbf{Datasets.} We conduct experiments on two widely used datasets, SYSU-MM01~\cite{hao2021cross} and RegDB~\cite{s17030605}.
The SYSU-MM01 dataset is collected from four visible and two infrared cameras. SYSU-MM01 provides a diverse range of indoor and outdoor environments, offering comprehensive scenarios. SYSU-MM01 consists of 287,628 visible images and 15,792 infrared images, with 491 individuals. The RegDB dataset comprises 412 individuals, each with ten visible and ten infrared images.

\noindent\textbf{Evaluation Metrics.} We conduct our experiments using standard evaluation settings employed in prior cross-modality Re-ID research~\cite{wu2023unsupervised}:   Mean Average Precision (mAP) and mean Inverse Negative Penalty (mINP). The SYSU-MM01 dataset includes two evaluation modes, all-search, and indoor-search. In the all-search mode, the gallery set consists of images from all visible cameras, while in the indoor-search mode, it only includes images from indoor visible cameras. On both modes, we conduct the experiments under single-shot setting~\cite{wang2022optimal}. The RegDB dataset has two testing modes, one with infrared images as the query set, visible images as the gallery set, and the other with reversed roles. We randomly divided the training and testing sets ten times and averaged the experiment results.

\noindent\textbf{Implementation details.}
The proposed method is implemented using PyTorch with two A6000 GPUs. AGW~\cite{ye2021deep} is utilized as the encoder to extract features.
The size of the input images is $288\times144$. The model optimization is based on the Adam Optimizer, which is initialized with a learning rate of 3.5e-3. At the 20th and 50th epochs, the learning rate decays by a factor of ten and is combined with a warm-up technique. We train the model for 75 epochs.

\noindent \textbf{Training Procedure}
We adopt a three-step training approach. Firstly, we train the encoder independently to extract discriminative features from inputs. Secondly, the generator and discriminator are trained jointly: the generator produces realistic features based on extracted features, while the discriminator learns to differentiate between real and generated features. This adversarial training enhances the generator's authenticity and the discriminator's discrimination ability. Finally, we fine-tune the entire model by jointly training the encoder, generator, and discriminator with $\mathcal{L}{plc}$ and $\mathcal{L}{cma}$ objectives. This comprehensive optimization approach improves individual components and promotes cooperation among strategies, resulting in significant performance enhancements.

\begin{table*}[t!]
\caption{Ablation study about loss function on SYSU-MM01 dataset.}
\renewcommand{\arraystretch}{1.15}
\centering
\captionsetup{font=small}

\resizebox{1\textwidth}{!}{
\begin{tabular}{ccccc||cccc||cccc}
\toprule
\multirow{2}*{ClusterNCE} &	\multirow{2}*{S-PLC}	& \multirow{2}*{D-PLC}	& \multirow{2}*{$\mathcal{L}_{cma}$}	 & \multirow{2}*{$\mathcal{L}_{lfc}$} & \multicolumn{4}{c||}{All Search} & \multicolumn{4}{c}{Indoor Search}\\ \cmidrule{6-13}
 & & & & & Rank-1 & Rank-5 & Rank-10 & mAP & Rank-1 & Rank-5 & Rank-10 & mAP \\ \midrule
\checkmark&	-&	- &- & - &39.39	&68.40	&79.80	&37.73	& 44.20	& 77.22	&87.09	& 54.31\\
-&	\checkmark&	-&	-	&- &41.32	& 70.13	& 81.56	&39.81	& 46.25	& 79.01	& 88.93	& 56.05\\
-&	-&	\checkmark&	-	&- &42.86	& 71.48	& 83.07	&41.11	& 47.93	& 80.57	& 90.15	& 57.26\\
\hline
\hline
-&	\checkmark&	-&	\checkmark	&- & 57.53	& 84.25	& 91.51	& 51.77	& 57.05	& 83.96	& 89.65	& 62.75\\
-&	\checkmark&	-&	-	&\checkmark & 56.82	& 83.51	& 90.63	&50.91	& 56.91	& 83.05	& 89.55	& 61.52\\
-&	\checkmark&	-&	\checkmark	&\checkmark &58.85	& 85.63	& 93.16	& 52.71	& 59.76	& 86.83	& 93.05	& 64.73\\
-&	-&	\checkmark&	\checkmark	&-	&58.35	& 84.74	& 92.17	&52.59	& 58.91	& 85.13	& 92.11	& 63.55\\
-&	-&	\checkmark	&-	&\checkmark	&58.21	& 84.52	& 91.83	&52.50	& 58.14	& 84.87	& 91.58	& 63.01\\
-&	-&	\checkmark&	\checkmark	&\checkmark	&59.44	& 85.87	& 93.78	&53.27	& 61.03	& 87.25	& 94.31	& 66.35\\
 \hline
\end{tabular}
}
\label{tab:Ablation two strategies}
\end{table*}

\subsection{Comparison  Experiments}
\label{Comparison  Experiments}

To demonstrate the effectiveness of our PRAISE, we compare it with three related ReID settings: supervised visible-infrared ReID (SVI-ReID) methods \ie
 DART~\cite{yang2022learning}, unsupervised visible-ReID (USL-ReID) \ie, PPLR~\cite{cho2022part}, and unsupervised visible-infrared ReID (USL-VI-ReID) \ie, ADCA~\cite{lin2019bottom}. The experiment results based on the SYSU-MM01 dataset and RegDB dataset are shown in Tab.~\ref{tab:SYSU Comparison} and Tab.~\ref{tab:RegDB Comparison}.

Compared to the SVI-ReID method, our PRAISE shows slightly lower mAP and mINP performance compared with current state-of-the-art (SOTA) methods. However, as the rank increases from Rank-1 to Rank-20, PRAISE demonstrates matching accuracy closer to SVI-ReID's, indicating the effectiveness of our approach in reducing errors in intra-modality clustering and cross-modality matching.

In contrast to USL-ReID methods trained directly on mixed datasets, our PRAISE outperforms SOTA methods across various evaluation metrics. This suggests that PRAISE can more effectively obtain identity-discriminative features for pedestrians across different modalities, resulting in higher accuracy.
Compared to USL-VI-ReID methods, PRAISE achieves SOTA performance across various evaluation metrics. This indicates that our proposed PLC and MLA strategy have alleviated issues related to noisy pseudo-labels and misalignment in scenarios with cross-modal data.

\begin{table}[t!]
\caption{Comparison with different matching methods on SYSU-MM01 dataset.}
\renewcommand{\arraystretch}{1.25}
\centering
\captionsetup{font=small}
\resizebox{0.49\textwidth}{!}{
\begin{tabular}{c||ccc||ccc}
\toprule
\multirow{2}*{Method} & \multicolumn{3}{c||}{All Search} & \multicolumn{3}{c}{Indoor Search} \\ \cmidrule{2-7}

 & Rank-1\% & mAP & mINP & Rank-1\% & mAP & mINP \\ \hline
GM+$\mathcal{L}_{cma}$ & 57.13	& 51.26	& 33.57	& 55.93	& 61.38	& 57.41 \\
GM+$\mathcal{L}_{cma}$+$\mathcal{L}_{lfc}$ & 58.12	& 52.35	& 35.48	& 57.65	& 63.38	& 59.37 \\
SFM+$\mathcal{L}_{cma}$  & 58.35	& 52.59	& 35.78	& 58.91	& 63.55	& 59.96 \\
SFM+$\mathcal{L}_{cma}$+$\mathcal{L}_{lfc}$	& 59.44	& 53.27	& 36.95	& 61.03	& 66.35	& 62.21 \\
 \hline
\end{tabular}
}

\label{tab:matching methods}
\end{table}

\begin{table}[t!]
\caption{Ablation about $\lambda_1$ and $\lambda_2$ on SYSU-MM01 dataset (‘all search’ mode).}
\centering
\captionsetup{font=small}
\resizebox{0.49\textwidth}{!}{
\begin{tabular}{c||ccc||ccc}
\toprule
 & \multicolumn{3}{c||}{mAP} & \multicolumn{3}{c}{mINP} \\ \hline

\diagbox{$\lambda_1$}{$\lambda_2$} & 0 & 0.5 & 1 & 0 & 0.5 & 1 \\ \hline
0 & 37.73	& 52.50	& 49.86	& 26.07	& 35.27	& 34.73 \\
0.5 & 52.59	& \textbf{53.27}	& 51.31	& 35.78	& \textbf{36.95}	& 35.54 \\
1 & 50.01	& 51.69	& 51.17	& 34.82	& 35.63	& 35.47 \\
 \hline
\end{tabular}
}

\label{tab:lamda}
\end{table}

\subsection{Ablation Study}
In this section, we conduct ablation studies to validate the effectiveness of the proposed pseudo-label correction strategy and modality-level alignment strategy.

\begin{figure}[t!]
    \centering
    \includegraphics[width=0.48\textwidth]{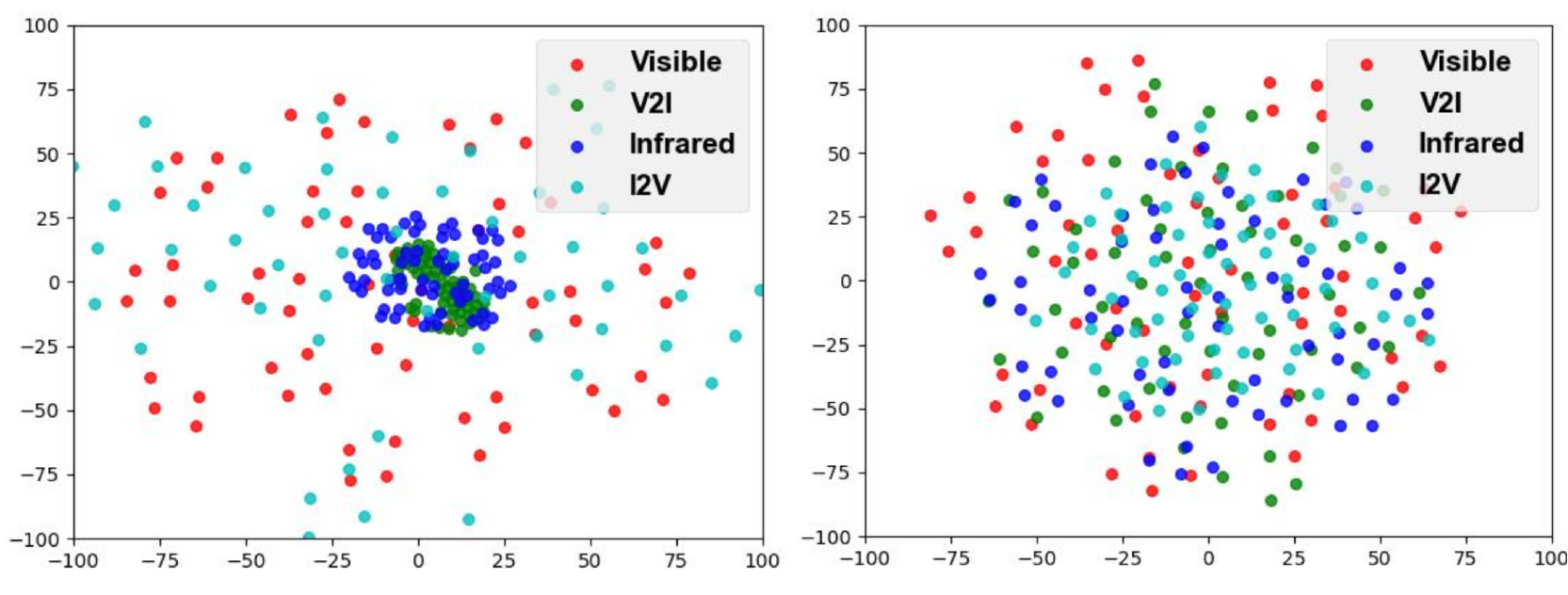}
    \caption{TSNE visualization of before MLA (left) and after MLA (right) in the same batch.}
    \vspace{-10pt}
    \label{Tsne_single}

\end{figure}

\begin{figure}[t!]
    \centering
    \includegraphics[width=0.48\textwidth]{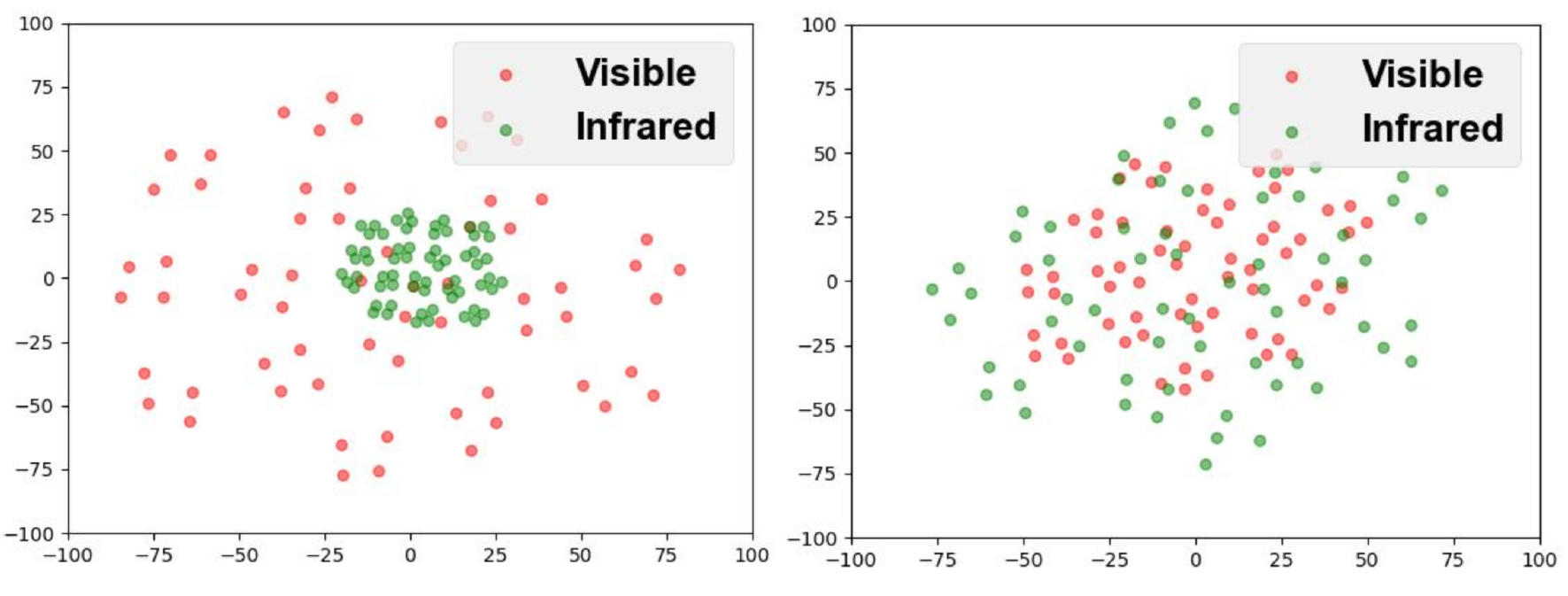}
    \caption{TSNE visualization of before MLA (left) and after MLA (right) of the same identity (randomly selected).}
    \label{Tsne}
    \vspace{-10pt}
\end{figure}
\noindent \textbf{Ablation studies on PLC strategy.} We evaluate the proposed PLC strategy ($\mathcal{L}_{plc}$) on the SYSU-MM01 dataset. The results are presented in Tab.~\ref{tab:Ablation two strategies}. The experiment results show that our PLC strategy yields performance improvements in mAP. Specifically, using ClusterNCE alone yields a Rank-1 accuracy of \textbf{39.39}\% in the 'all search' scenario and \textbf{44.20}\% in the indoor search scenario. Similarly, using S-PLC or D-PLC alone leads to improved performance, with Rank-1 accuracies of \textbf{41.32}\% and \textbf{42.86}\% respectively in all search scenarios, and \textbf{46.25}\% and \textbf{47.93}\% respectively in the indoor search scenario.

\noindent \textbf{Ablation studies on $\mathcal{L}_{cma}$ and $\mathcal{L}_{lfc}$.}  We conduct several experiments to validate the effectiveness of the proposed $\mathcal{L}_{cma}$ and $\mathcal{L}_{lfc}$. The results of these experiments are presented in Tab.~\ref{tab:Ablation two strategies}. Initially, we employed S-PLC for contrastive learning within the modality. By incorporating $\mathcal{L}_{cma}$ and $\mathcal{L}_{lfc}$, we observed enhancements in the model's recognition performance. Specifically, utilizing the  $\mathcal{L}_{cma}$ can improve the performance with a \textbf{9.95} mAP improvement, while the $\mathcal{L}_{lfc}$ can lead to a \textbf{9.07} mAP enhancement. In addition, we also used D-PLC to verify the performance of $\mathcal{L}_{cma}$ and $\mathcal{L}_{lfc}$. After adding $\mathcal{L}_{cma}$ and $\mathcal{L}_{lfc}$, the model performance improved by \textbf{9.1} mAP and \textbf{8.76} mAP respectively. When two losses are introduced, the model achieves the best performance. The ablation of the hyper-parameters $\lambda_1$ and $\lambda_2$ are shown in Tab.~\ref{tab:lamda}.

\noindent \textbf{Effectiveness of SFM.} Tab.~\ref{tab:matching methods} presents a comparison of different matching methods to evaluate the performance of the SFM. The SFM method outperforms the graph marching (GM)~\cite{wu2023unsupervised} method in all search scenarios. Specifically, SFM achieves a Rank-1 accuracy of \textbf{58.35}\% and \textbf{58.91}\% for all search and indoor search respectively, while GM achieves \textbf{57.13}\% and \textbf{55.93}\% in the same scenarios. Similarly, SFM achieves higher mAP values of \textbf{52.59}\% and \textbf{63.55}\% for all search and indoor search respectively, compared to GM's mAP values of \textbf{51.26}\% and \textbf{61.38}\%. The results also show that the proposed functions further improve the performance of both GM and SFM.

\subsection{Discussion}
\noindent \textbf{Visualization about MLA.}To intuitively observe the impact of reducing the modal disparity on the spatial distribution, we employ the t-SNE~\cite{van2008visualizing} to visualize the distribution of the two modalities in a two-dimensional space, as shown in Fig~\ref{Tsne_single} and Fig.~\ref{Tsne}.
Fig.~\ref{Tsne_single} shows that using the same data batch reduces the modality gap between visible and infrared, while employing CMA loss increases distances between identities, resulting in a more concentrated distribution in a smaller space. This compression enhances cross-modality alignment and diminishes modality gaps.
Fig.~\ref{Tsne} visually shows results before and after applying the MLA strategy, using identical identity data. The MLA strategy reduces the modality gap within the same identity and aligns cross-modality data effectively. These visualizations further support the effectiveness of the MLA strategy in addressing modality gap issues.
\label{Visualization}


\section{Conclusion}

In this paper, we proposed a fully unsupervised framework for VI-ReID. Specifically, we proposed two novel strategies to enhance the performance of VI-ReID by mitigating the detrimental effect of noisy pseudo labels and reducing the modality gap. Specifically, we introduced the pseudo label correction and modality-level alignment strategies to let the model focus on identity-discriminative and modality-invariant features, respectively.
The experimental results robustly validate the effectiveness of our proposed strategies, showcasing a substantial improvement in the model's matching performance. Additionally, the proposed two strategies can be easily injected into further cross-modal representation learning methods.

\noindent \textbf{Future work:}
We employ CycleGAN for cross-modal feature translation, but the current three-stage training process is complex for stability. In future work, we aim to design a simplified training method. Besides, the unlabeled visible and infrared images are still used for training; therefore, we plan to investigate the source-free method in our future work, which can avoid the exposure of visible information and might mitigate security risks.
{
    \small
    \bibliographystyle{IEEEtran}
    \bibliography{citation}
}

\end{document}